%% file: repcolt.tex
\documentclass{article}
\usepackage{amsmath}
\usepackage{amsthm}
\usepackage{amssymb}
\usepackage{mathtools}
\usepackage{epsf}
\usepackage{colt95}

\input{mathdefs}

\newenvironment{pf}{\noindent{\bf Proof.\ }}{$\Box$ \\ }

\begin{document}

\title{Learning Internal Representations (COLT 1995)}
\author{Jonathan Baxter\thanks{This work was supported by a Shell Australia Postgraduate Fellowship and an Australian Postgraduate Research Award.}
\\
  Department of Mathematics and Statistics\\
  The Flinders University of South Australia}
\maketitle
{\sloppy
\begin{abstract}
Probably the most important problem in machine learning is the preliminary
biasing of a learner's hypothesis space so that it is small enough to ensure
good generalisation from reasonable training sets, yet large
enough that it contains a good solution to the problem being learnt.
In this paper a mechanism for {\em automatically} learning or
biasing the learner's hypothesis space is introduced. 
It works by first learning an appropriate {\em internal representation} for
a learning environment and then using that representation to bias the learner's
hypothesis space for the learning of future tasks drawn from the same
environment.

An internal representation must be learnt by sampling from {\em many similar
tasks}, not just a single task as occurs in
ordinary machine learning. It is proved that the
number of examples $m$ {\em per task} required to ensure good generalisation
from a representation learner obeys $m = O(a+b/n)$ where $n$ is the number
of tasks being learnt and $a$ and $b$ are constants. If the tasks are learnt
independently ({\em i.e.} without a common representation) then $m=O(a+b)$.
It is argued that for
learning environments such as speech and character recognition $b\gg a$ and
hence representation learning in these environments can potentially yield a drastic
reduction in the number of examples required per task. It is also
proved that if $n = O(b)$ (with $m=O(a+b/n)$) 
then the representation learnt will be good for
learning novel tasks from the same environment, and that the number of
examples required to generalise well on a novel task will be reduced to $O(a)$
(as opposed to $O(a+b)$ if no representation is used). 

It is shown that gradient descent can be used to train neural network
representations and the results of an experiment are reported in which a
neural network representation was learnt for an environment consisting
of {\em translationally invariant} Boolean functions. The experiment
provides strong qualitative support for the theoretical results. 
\end{abstract}

\section{Introduction}
It has been argued elsewhere (for example, see \cite{Getal}) that the main
problem in machine learning is the biasing of the learner's hypothesis space
sufficiently well to ensure good generalisation from a relatively small
number of examples. Once suitable biases have been found the actual learning
task is relatively trivial. Despite this conclusion, much of machine
learning theory is still concerned only with the problem of quantifying the
conditions necessary for good generalisation {\em once} a suitable
hypothesis space for the learner has been found; virtually no work appears
to have been done on the problem of how the learner's hypothesis space is to
be selected in the first place. This paper presents a new method 
for automatically selecting a learner's
hypothesis space: {\em internal representation learning}.

The idea of automatically learning internal representations is not new to
machine learning. In fact the huge increase in Artificial Neural Network
(henceforth ANN) research over the last decade can be partially attributed
to the promise---first given air in \cite{Retal}---that neural networks can
be used to {\em automatically} learn appropriate internal representations.
However it is fair to say that despite some notable isolated successes,
ANNs have failed to live up to this early promise.
The main reason for this is not any inherent deficiency of ANNs as a machine
learning model, but a failure to realise the true source of information
necessary to generate a good internal representation. 

Most machine learning
theory and practice is concerned with learning a {\em single} task (such as
``recognise the digit `1'\,'') or at most a handful of tasks (recognise the
digits `0' to `9'). However it is unlikely that the information contained
in a small number of tasks is sufficient to determine an appropriate
internal representation for the tasks. To see this,
consider the problem of learning to recognise the
handwritten digit `1'.
The most extreme representation possible 
would be one that completely solves the classification problem,
i.e. a representation that outputs `yes' if its input is an image of a
`1', regardless of the 
position, orientation, noise or writer dependence of the original digit, and
`no' if any other image is presented to it. A learner equipped with such a
representation would require only one positive and one negative example of
of the digit to learn to recognise it perfectly.
Although the representation in this example certainly reduces the complexity
of the learning problem, it does not really seem to capture what is meant by the term
{\em representation}. What is wrong is that although the representation is an
excellent one for learning to recognise the digit `1', 
{\em it could not be used for any
other learning task}. A representation that is appropriate for learning to
recognise `1' should also be appropriate for other character recognition
problems---it should be good for learning other digits, 
or the letters of the alphabet, or Kanji
characters, or Arabic letters, and so on. Thus the information necessary
to determine a good representation is not contained in a single learning
problem (recognising `1'), but is contained in many examples of similar
learning problems. The same argument can be applied to other familiar
learning domains, such as face recognition and speech recognition. A
representation appropriate for learning a single face should be appropriate
for learning all faces, and similarly a single word representation should be
good for all words (to some extent even regardless of the language). 

In the rest of this paper it is shown how to formally model the process
of sampling from many similar learning problems and 
how information from such a sample can be used to learn an appropriate
representation. If $n$ learning problems are learnt independently then the
number of examples $m$ required per problem for good generalisation 
obeys $m=O(a+b)$, whereas if a
common representation is learnt for all the problems then $m=O(a+b/n)$. The
origin of the constants $a$ and $b$ will be explained in section
\ref{repsec} and it is argued in section \ref{mathsec}
that for common learning domains such as
speech and image recognition $b\gg a$, hence representation learning in
such environments can potentially yield a drastic reduction in the number of
examples required for good generalisation. It will also be shown that 
if a representation is learnt on $n=O(b)$ tasks then with high probability
it will be good for learning novel tasks and that the 
sampling burden for good generalisation on novel tasks will be
reduced $m=O(a)$, in constrast to
$m=O(a+b)$ if no representation is used.

In section \ref{backpropsec} it is shown how to use gradient descent to train
neural network representations, and the results of an experiment using the
technique are reported in section \ref{expsec}. The experiment
provides strong qualitative support for the theoretical results.

\section{Mathematical framework}
\label{mathsec}
Haussler's \cite{Haussler} {\em statistical decision theoretic} formulation
of ordinary machine learning is used throughout this paper as it has the widest
applicability of any formulation to date. This formulation may be summarised
as follows.
The learner is provided with a {\em training set} $\zv = (z_1,\dots,z_m)$
where each {\em example} $z_i = (x_i,y_i)$ consists of an {\em input}
$x_i\in X$ and an {\em outcome} $y_i\in Y$. The training set is generated by
$m$ independent trials according to some (unknown) joint
probability distribution $P$ on $Z=X\times Y$. 
In addition the learner 
is provided with an {\it action} space $A$, a {\em loss function}
$l\colon Y\times A \to [0,M]$
and a {\em hypothesis
space} $\H$ containing functions $h\colon X\to A$.
Defining the 
{\em true loss}
of hypothesis $h$ with respect to distribution $P$ as
\begin{equation}
\label{terr}
E(h,P) \de \int\limits_\XY l\(y, h(x)\)\,dP(x,y),
\end{equation}
the goal of the learner
is to produce a hypothesis $h\in\H$ that has true loss as small as
possible.
$l(y,h(x))$ is designed to give a measure of the {\em loss} the learner
suffers, when, given an input $x\in X$, it produces an action $h(x)\in A$ and
is subsequently shown the outcome $y\in Y$.

If for each $h\in \H$ a function $l_h\colon Z\to[0,M]$ is defined by 
$l_h(z) = l(y,h(x))$ for all 
$z=(x,y)\in Z$, then $E(h,P)$ can be expressed as 
the expectation of 
$l_h$ with respect to $P$,
$$
E(h,P) = \E{l_h}{P} = \int_Z l_h(z)\, dP(z).
$$
Let $l_\H = \{l_h\colon h\in\H\}$.
The measure $P$ and the 
$\sigma$-algebra on $Z$ are assumed to be such that all the $l_h\in l_\H$ 
are $P$-measurable (see definition \ref{measdef}).

To minimize the true loss the learner 
searches for a hypothesis 
minimizing the {\em empirical loss} on the sample $\zv$,
\begin{equation}
\label{eerr}
E(h,\zv) = \E{l_h}{\zv} = \frac1m\sum_{i=1}^m l_h(z_i).
\end{equation}

To enable the learner to
get some idea of the environment in which it is learning and hopefully then
extract some of the bias inherent in the environment,
it is assumed that the environment consists of a {\em set} of probability
measures $\P$ and an {\em environmental measure} $Q$ on $\P$.
Now the learner is not just supplied
with a single sample $\zv =(\pr{z}{,}{m})$, sampled according to some
probability measure $P\!\in\!\P$, but with $n$ such samples
$\(\pr{\zv}{,}{n}\)$. Each sample $\zv_i=\(z_{i1},\dots,z_{im}\)$, for
$\iton$, is generated by first sampling from $\P$ according to the
environmental measure $Q$ to
generate $P_i$, and then sampling $m$ times from $P_i$ to generate
$\zv_i=\(z_{i1},\dots,z_{im}\)$.
Denote the entire sample by 
$\z$
and write it as an $n\times m$ ($n$ rows, $m$ columns) matrix over $Z$:
$$
\z =
\begin{array}{ccc}
z_{11} & \cdots & z_{1m} \\
\vdots & \ddots & \vdots \\
z_{n1} & \cdots & z_{nm} 
\end{array}
$$
Denote the $n\times m$ matrices over $Z$ by $\Znm$ and call a sample
$\z\in\Znm$ generated by the above process an {\em $(n,m)$-sample}.

To illustrate this formalism, consider the problem of {\em character
recognition}. In this case $X$ would be the space of all images, $Y$ would
be the set $\{0,1\}$, each probability measure $P\in\P$ would represent a
distinct character or character like object in the sense that $P(y|x)=1$
if $x$ is an image of the character $P$ represents and
$y=1$, and $P(y|x) = 0$ otherwise. The marginal distribution 
$P(x)$ in each case could be formed by choosing a positive example of the
character with probability half from the `background'
distribution over images of the character concerned, and similarly 
choosing a negative
example with probability half. $Q$ would give the probability of
occurence of each character. The $(n,m)$-sample $\z$ is then simply a set of
$n$ training sets, each row of $\z$ being a sequence of $m$ classified examples
of some character.
If the idea of character is widened to include
other alphabets such as the greek alphabet and the Japanese Kanji
characters then the number of different characters to sample from is very
large indeed.

To enable the learner to take advantage of
the prior information contained in the $(n,m)$-sample $\z$,
the hypothesis space $\H\colon X\to A$ is split into 
two sections: $\H = \comp{\G}{\F}$ where $\F\colon X\to V$ and $\G\colon V\to A$, where 
$V$ is an arbitrary set\footnote{That is, $\H=\{\comp{g}{f}\colon g\in\G,
f\in\F\}$.}. 
To simplify the notation this will be written in future as  
$$
\XVA.
$$
$\F$ will be called the {\em representation space}
and an individual member $f$ of $\F$ will be  
called an {\em internal representation} or just a {\em representation}.
$\G$ will be called the {\em output function space}.

Based on the information about the environment $Q$, contained in $\z$, the
learner searches for a good representation $f\in\F$. A good representation is 
one with a small {\em empirical loss}
$E^*_\G(f,\z)$ on $\z$, where this is defined by
\begin{equation}
\label{emploss}
E^*_\G(f,\z) \de \frac1n\sum_{i=1}^n \inf_{g\in\G}\E{l_\comp{g}{f}}{\zv_i},
\end{equation}
where $\zv_i = \(z_{i1},\dots,z_{im}\)$ denotes the $i$th row of $\z$.
The empirical loss of $f$ with respect to $\z\in\Znm$ is a 
measure of how well the learner can learn $\z$ 
using $f$, assuming that the learner 
is able to find the best possible $g\in\G$ for any given sample $\zv\in\Zm$. 
For example, if the empirical loss
of $f$ on $\z=(\zv_1,\dots,\zv_n)$ is zero then it is possible for the 
learner to find an output function\footnote{Assuming the infimum is attained in $\G$.} 
$g_i\in\G$, for each $i$, $1\leq i\leq n$, such that the ordinary empirical
loss $\E{l_\comp{g_i}{f}}{\zv_i}$ is zero.
The empirical
loss of $f$ is an estimate of the {\em true loss} of $f$
(with respect to $Q$), where this is defined by
\begin{equation}
\label{exploss}
E^*_\G(f,Q)
\de \int_\P \inf_{g\in\G} \E{l_\comp{g}{f}}{P}\, dQ(P).
\end{equation}
The true loss of $f$ with respect to $Q$ is the expected best possible
performance of $\comp{g}{f}$---over all 
$g\in\G$---on a distribution chosen at random
from $\P$ according to $Q$. If $f$ has a small true loss then learning using
$f$ on a random ``task'' $P$---drawn according to $Q$---will with high
probability be successful. 

Note that the learner takes as input 
samples $\z\in\Znm$, for any values $n,m\ge 1$, and produces as output
hypothesis representations $f\in\F$, so it is a map $\A$ from the
space of all possible $(n,m)$ samples into $\F$,
$$
\A\colon  \bigcup_{n,m\geq 1} \Znm \to \F.
$$

It may be that the $n$ tasks $\Pv=(P_1,\dots,P_n)$ generating the training 
set $\z$ are all that the
learner is ever going to be required to learn, in which case it is more
appropriate to suppose that in response to the $(n,m)$-sample $\z$ the
learner generates $n$ hypotheses 
$\comp{\gv}{\fbar} = (\comp{g_1}{f}, \dots, \comp{g_n}{f})$ all using the
same representation $f$ and collectively minimizing
\begin{equation}
\label{egnferr}
E(\comp{\gv}{\fbar},\z) = \frac1n\sum_{i=1}^n \E{l_{\comp{g_i}{f}}}{\zv_i}.
\end{equation}
The true loss of the learner will then be 
\begin{equation}
\label{tgnferr}
E(\comp{\gv}{\fbar},\Pv) = \frac1n\sum_{i=1}^n \E{l_{\comp{g_i}{f}}}{P_i}.
\end{equation}
\sloppy
Denoting the set of all functions $\comp{\gv}{\fbar} =
(\comp{g_1}{f},\dots,\comp{g_n}{f})$ for $g_1,\dots,g_n\in\G$ and $f\in\F$ 
by $\comp{\G^n}{\Fbar}$, the learner in this case is a map
$$
\A\colon  \bigcup_{n,m\geq 1} \Znm \to \comp{\G^n}{\Fbar}.
$$

If the learner is going to be using the representation $f$ to learn future
tasks drawn according to the same environment $Q$, it will do so by 
using the {\em restricted} 
hypothesis space $\comp{\G}{f} \de \{\comp{g}{f}\colon g\in\G\}$. That is,
the learner will be fed samples $\zv\in\Zm$ drawn according to some
distribution $P\in\P$, which in turn is drawn according to $Q$, and will
search $\comp{\G}{f}$ for a hypothesis $\comp{g}{f}$ with small empirical
loss on $\zv$. 
Intuitively, if $\F$ is much ``bigger'' than 
$\G$ then the number of examples required to learn using $\comp{\G}{f}$ will
be much less than the number of examples required to learn using the full
space $\comp{\G}{\F}$, a fact that is proved in the next section. Hence,
if the learner can find a good representation $f$ and the sample $\z$ is
large enough, learning using $\comp{\G}{f}$ will be considerably quicker
and more reliable than learning using $\comp{\G}{\F}$. If the representation
mechanism outlined here is the one employed by our brains then the fact that
children learn to recognise characters and words from a relatively small
number of examples is evidence of a small $\G$ in these cases. The fact that
we are able to recognise human faces after being shown only a single example
is evidence of an even smaller $\G$ for this task. Furthermore, the fact
that most of the difficulty in machine learning lies in the initial bias of
the learner's hypothesis space \cite{Getal} indicates that our ignorance
concerning an appropriate representation $f$ is large, and hence the entire
representation space $\F$ will have to be large to ensure that it contains a
suitable representation. Thus it seems that at least for the examples
outlined above the conditions
ensuring that representation learning is a big improvement over ordinary
learning will be satisfied.

The main issue in machine learning is that of quantifying the necessary
sampling conditions ensuring good generalisation. In representation
learning there are two measures of good generalisation. The first is the
proximity of \eqref{egnferr} above to the second form of
the true loss \eqref{tgnferr}. If the sample $\z$ is large enough to
guarantee with high probability that these two quantities are close, then a
learner that produces a good performance in training on $\z$ will be likely
to perform well on future examples of any of the $n$ tasks used to generate
$\z$. 
The second measure of generalisation performance is the proximity of
\eqref{emploss} to the first form of the true loss \eqref{exploss}. In this
case good generalisation means that the learner should expect to perform
well if it uses the representation $f$ to learn a 
new task $P$ drawn at random according to the
environmental measure $Q$. 
Note that this is a new form of generalisation, one
level of abstraction higher than the usual meaning of generalisation, for
within this framework a
learner generalises well if, after having learnt many different tasks,
it is able to learn new tasks easily. Thus, not only is the learner 
required to generalise well in the ordinary sense by generalising well on
the tasks in the training set,
but also the learner is expected to have ``learnt to learn'' the tasks from
the environment in general. Both the number of tasks $n$ generating $\z$ and
the number of examples $m$ of each task must be sufficiently large to ensure
good generalisation in this new sense. 

To measure the deviation between \eqref{egnferr} and \eqref{tgnferr}, and 
\eqref{emploss} and \eqref{exploss}, the following 
one-parameter family of metrics
on $\R^+$, introduced in \cite{Haussler}, will be used:
$$
d_\nu(x,y) = \frac{|x-y|}{\nu+x+y},
$$
for all $\nu>0$ and $x,y\in\R^+$.
Thus, good generalisation in the first case is governed by the probability
\begin{equation}
\label{p1}
\Pr\left\{\z\in\Znm:\d{E(\A(\z),\z)}{E(\A(\z),\Pv)}
> \alpha\right\},
\end{equation}
where the probability measure on $\Znm$ is $P_1^m\times\dots
\times P_n^m$. 
In the second case it is governed by the probability 
\begin{equation}
\label{p2}
\Pr\left\{\z\in\Znm:\d{E^*_\G(\A(\z),\z)}{E^*_\G(\A(\z),Q)} >
\alpha\right\}. 
\end{equation}
This time the probability measure on $\Znm$ is
$$
\mu(S) = \int_{\P^n} P_1^m\times\dots\times P_n^m(S)\,
dQ^n(P_1,\dots,P_n)
$$
for any measurable subset\footnote{The minimal $\sigma$-algebra for $Z$ is
$\sigma_{l_\comp{\G}{\F}}$, as defined in definition \ref{measdef}.}
$S$ of $\Znm$.

Conditions on the sample $\z$ ensuring the probabilities \eqref{p1} and
\eqref{p2} are small are derived in the next section.

\section{Conditions for good generalisation}
\label{repsec}
To state the main results some further definitions are required.
\begin{defn}
\label{metricdef}
{\rm
For the structure $\XVA$ and loss function $l\colon Y\times A\to [0,M]$
define $l_g\colon V\times Y\to [0,M]$ for any $g\in\G$ by $l_g(v,y) =
l(y,g(v))$. Let $l_\G=\{l_g\colon g\in\G\}$.
For any probability measure $P$ on $V\times Y$ 
define the pseudo-metric $d_P$ on $l_\G$ by
\begin{equation}
\label{pmetric}
d_P(l_g,l_{g'}) = \int_{V\times Y} |l_g(v,y)-l_{g'}(v,y)|\,dP(v,y).
\end{equation}
Let $\N(\ep,l_\G,d_P)$ be the size of the smallest $\ep$-cover of $(l_\G,d_P)$
and define the {\em $\ep$-capacity} of $l_\G$ to be 
\begin{equation}
\label{capacity}
\C(\ep,l_\G) = \sup \N(\ep,l_\G,d_P)
\end{equation}
where the supremum is over all probability measures $P$.
For any probability measure $P$ on $Z$
define the pseudo-metric $d^*_{[P,l_\G]}$ on $\F$ by\footnote{For
$d^*_{[P,l_\G]}$ to be well defined the supremum 
over $\G$ must be $P$-measurable. This is guaranteed if the hypothesis space
{\em family} $\{l_\comp{\G}{f}\colon f\in\F\}$ is {\em f-permissible} (see
appendix \ref{permapp}).}
$$
d^*_{[P,l_\G]}(f,f') = \int_Z\sup_{g\in\G} |l_\comp{g}{f}(z) -
l_\comp{g}{f'}(z)|\,dP(z). 
$$
Let $\C^*_{l_\G}(\ep,\F)$ be the corresponding $\ep$-capacity.
}
\end{defn}
\subsection{Generalisation on $\boldmath{n}$ tasks}
The following theorem bounds the number of examples $m$ of each task in an
$(n,m)$-sample $\z$ needed to ensure with high probability 
good generalisation from a representation
learner on average on all the tasks. It uses the notion of a {\em hypothesis
space family}---which is just a set of hypothesis spaces---and a
generalisation of Pollard's concept of {\em permissibility} \cite{Pollard}
to cover hypothesis space families, called {\em f-permissibility}. The
definition of f-permissibility is given in appendix \ref{permapp}.
\begin{thm}
\label{thm1}
Let $\F$ and $\G$ be families of functions with the structure  $\XVA$,
let $l$ be a loss function $l\colon Y\times A\to [0,M]$ and suppose $\F,\G$
and $l$ are such that the
hypothesis space family $\{l_{\comp{\G}{f}}\colon f\in\F\}$ is
f-permissible. Let
$P_1,\dots,P_n$ be $n$ probability measures on $Z=X\times Y$ and let
$\z\in\Znm$ be an $(n,m)$-sample generated by sampling $m$ times from $Z$
according to each $P_i$. For all $0<\alpha<1, 0<\delta<1, \nu>0$ and any
representation learner 
$$
\A\colon  \bigcup_{n,m\geq 1}\Znm \to \comp{\G^n}{\Fbar},
$$
if 
\begin{equation}
\label{mbound1}
m\geq \frac{8M}{\alpha^2\nu}\left[\ln\C(\ep_1,l_\G) + 
\frac1n\ln\frac{4\C^*_{l_\G}(\ep_2,\F)}{\delta}\right]
\end{equation}
where $\ep_1+\ep_2 = \frac{\alpha\nu}{8}$, then
$$
\Pr\left\{\z\in\Znm\colon\d{E(\A(\z),\z)}{E(\A(\z),\Pv)} 
> \alpha\right\} \leq \delta.
$$
\end{thm}
\begin{pf}
See appendix \ref{proofapp}.
\end{pf}

Theorem \ref{thm1} with $n=1$ corresponds to the ordinary
learning scenario in which a single task is learnt. Setting
$a=\frac{8M}{\alpha^2\nu}\ln\C(\ep_1,l_\G)$ and 
$b=\frac{8M}{\alpha^2\nu}\ln\frac{4\C^*_{l_\G}(\ep_2,\F)}{\delta}$ 
gives $m=O(a+b)$ for a single task while $m=O(a+b/n)$ for $n$ tasks learnt
using a common representation\footnote{The choice of $\ep_1$ and $\ep_1$
subject to $\ep_1+\ep_2 = \alpha\nu/8$ 
giving the best bound on $m$ will differ between the $n=1$ and $n>1$ cases,
so strictly speaking {\em at worst} $m=O(a+b/n)$ in the
latter case.}. Note
also that if the $n$ tasks are learnt independently then
the learner is a map
from the space of all $(n,m)$-samples into $\comp{\G^n}{\F^n}$ rather than
$\comp{\G^n}{\Fbar}$ and so (not surprisingly) the number of examples $m$
per task required will be the same as in the single task case: $m=O(a+b)$.
Thus for hypothesis spaces in which $b\gg a$ learning $n$ tasks using a
representation will be far easier than learning them independently.

If $\F$ and $\G$ are Lipschitz bounded 
neural networks with $W_\F$ weights in $\F$ and $W_\G$ weights in $\G$
and $l$ is one of many loss functions
used in practice (such as Euclidean loss, mean squared loss, cross entropy
loss---see \cite{Haussler}, section 7), then simple extensions of
the methods of \cite{Haussler} can be used to show:
\begin{eqnarray*}
\C(\ep,l_\G) &\leq& \left[\frac{k}{\ep}\right]^{2W_\G} \\
\C^*_{l_\G}(\ep,\F) &\leq& \left[\frac{k'}{\ep}\right]^{2W_\F}
\end{eqnarray*}
where $k$ and $k'$ are constants not dependent on the number of weights or
$\ep$. Substituting these expressions into \eqref{mbound1} and optimizing
the bound with respect to $\ep_1+\ep_2=\alpha\nu/8$ gives:
\begin{eqnarray*}
\ep_1 &=& \frac{\min(W_\F,nW_\G)}{W_\F+nW_G}\frac{\alpha\nu}{8} \\
\ep_2 &=& \frac{\max(W_\F,nW_\G)}{W_\F+nW_\G}\frac{\alpha\nu}{8}
\end{eqnarray*}
which yields\footnote{Bound \eqref{nnetbound1} does not contradict known results on
the VC dimension of {\em threshold} neural networks of $W\log W$ because it
only applies to Lipschitz bounded neural networks, and the Lipschitz bounds
are part of the bound on $m$ (but they are not shown).}
\begin{equation}
\label{nnetbound1}
m = O\left( \frac1{\alpha^2\nu}\left[ \ln\frac1{\alpha\nu}\left(
W_\G+\frac{W_\F}{n} \right) + \frac1n\ln\frac1\delta \right] \right).
\end{equation}
Although \eqref{nnetbound1} results from a worst-case analysis and some
rather crude approximations to the capacities of $\F$ and $\G$, its general
{\em form} is intuitively very appealing---particularly the behaviour of $m$ 
as a function of $n$ and the size of $\F$ and $\G$. I would expect this
general behaviour to survive in more accurate analyses of specific
representation learning scenarios. This conclusion is certainly supported by
the experiments of section \ref{expsec}.

\subsection{Generalisation when learning to learn}
The following theorem bounds the number of tasks $n$ and the number of
examples $m$ of each task required of an $(n,m)$-sample $\z$ to ensure with high
probability that a representation learnt on that sample will be good for
learning future tasks drawn according to the same environment.

\begin{thm}
\label{thm2}
Let $\F, \G$ and $l$ be as in theorem \ref{thm1}.
Let $\z\in \Znm$ be an $(n,m)$-sample from $Z$ according 
to an environmental measure $Q$. 
For all $0<\alpha,\delta,\ep_1,\ep_2<1$, $\nu>0$, 
$\ep_1+\ep_2=\frac{\alpha\nu}{16}$, 
if 
\begin{eqnarray*}
n  &\geq& \frac{32 M}{\alpha^2\nu} 
\ln\frac{8 \C^*_{l_\G}\(\frac{\alpha\nu}{16}, \F\)}{\delta}, \\
\mbox{and} \quad m &\geq& \frac{32M}{\alpha^2\nu}\left[\ln\C\(\ep_1,l_\G\) +
\frac1n\ln\frac{8\C^*_{l_\G}\(\ep_2,\F\)}{\delta}\right],
\end{eqnarray*}
and $\A$ is any representation learner,
$$
\A\colon  \bigcup_{n,m\geq 1}\Znm \to \F,
$$
then
$$
\Pr\left\{\z\in \Znm\colon \d{E^*_\G(\A(\z),\z)}{E^*_\G(\A(\z),Q)} >
\alpha\right\} \leq \delta.
$$
\end{thm}
\begin{pf}
See appendix \ref{proofapp}.
\end{pf}

Apart from the odd factor of two, the bound on $m$ in this theorem is the
same as the bound in theorem \ref{thm1}. Using the notation introduced
following theorem \ref{thm1}, the bound on
$n$ is $n=O(b)$, which is very large. However it results again from a
worst-case analysis (it corresponds to the worst possible environment
$Q$) and is only an approximation, so it is likely to be beaten in practice. 
The experimental
results of section \ref{expsec} verify this.
The bound on $m$ now becomes $m=O(a)$, while the {\em total} number of
examples $mn = O(ab)$. This is far greater than the $O(a+b)$ examples that
would be required to learn a single task using the full space
$\comp{\G}{\F}$, however a representation learnt on a single task cannot be
used to reliably learn novel tasks. Also the representation learning phase
is most likely to be an {\em off-line} process to generate the preliminary
bias for later {\em on-line} learning of novel tasks, and as learning of novel
tasks will be done with the biased hypothesis space $\comp{\G}{f}$ rather than
the full space $\comp{\G}{\F}$, the number of examples required for good
generalisation on novel tasks will be reduced to $O(a)$. 

\section{Representation learning via backpropagation}
\label{backpropsec}
If the function classes $\F$ and $\G$ consist of ``backpropagation'' type 
neural networks then it is possible to use a slight variation on ordinary
gradient descent procedures to learn a neural network representation $f$.

In this case, based on an $(n,m)$-sample 
$$
\z =
\begin{array}{cccc}
(x_{11},y_{11}) & (x_{12}, y_{12}) &\cdots & (x_{1m},y_{1m})\\ 
(x_{21},y_{21}) & (x_{22},y_{22})  &\cdots & (x_{2m},y_{2m})\\ 
\vdots & \vdots & \ddots & \vdots \\
(x_{n1},y_{n1}) & (x_{n2},y_{n2}) &\cdots & (x_{nm},y_{nm}),
\end{array}
$$
the learner searches for a representation $f\in\F$ minimizing the 
{\em mean-squared} representation error:
$$
E^*_\G(f,\z) = \frac1n\sum_{i=1}^n \inf_{g\in\G} 
\frac{1}{m}\sum_{j=1}^m \(\comp{g}{f}(x_{ij}) - y_{ij}\)^2.
$$
The most common procedure for training differentiable neural 
networks is to use some form of gradient descent algorithm (vanilla backprop,
conjugate gradient, etc) to minimize the error of the network on the sample
being learnt. For example, in ordinary learning the learner would receive
a single sample $\zv = ((x_1,y_1),\dots,(x_m,y_m))$ and would 
perform some form
of gradient descent to find a function $\comp{g}{f}\in\comp{\G}{\F}$ such that
\begin{equation}
\label{orderr}
E(\comp{g}{f},\zv) = \frac1m\sum_{i=1}^m\(\comp{g}{f}(x_i)-y_i\)^2
\end{equation}
is minimal. This procedure works because it is a relatively simple matter 
to compute the gradient, $\nabla_\w E(\comp{g}{f},\z)$, where $\w$
are the parameters (weights) of the networks in $\F$ and $\G$. 

Applying this
method directly to the problem of minimising $E^*_\G(f,\z)$ above would mean
calculating the gradient $\nabla_\w E^*_\G(f,\z)$ where now $\w$
refers only to the parameters of $\F$. However, due to the presence of the
infimum over $\G$ in the formula for $E^*_\G(f,\z)$, calculating 
the gradient in this case is much more difficult than in the ordinary
learning scenario. An easier way to proceed is to minimize
$E(\comp{\gv}{\fbar},\z)$ 
(recall equation \eqref{egnferr})
over all $\comp{\gv}{\fbar}$, for if
$\comp{\gv}{\fbar}$ is such that 
$$
\left|E(\comp{\gv}{\fbar},\z) - 
\inf_{\comp{\gv}{\fbar}\in\comp{\G^n}{\Fbar}} E(\comp{\gv}{\fbar},\z)\right|
\leq \ep, 
$$
then so too is
$$
\left|E^*_\G(f,\z) - \inf_{f\in\F} E^*_\G(f,\z)\right| \leq \ep.
$$
The advantage of this approach is that essentially the 
same techniques used for computing the gradient in ordinary learning can 
be used to compute the gradient of $E(\comp{\gv}{\fbar},\z)$.
Note that in the present framework $E(\comp{\gv}{\fbar},\z)$ is:
\begin{equation}
\label{gnferr}
E(\comp{\gv}{\fbar},\z) = \frac1n\sum_{i=1}^n\frac1m\sum_{j=1}^m
\(\comp{g_i}{f}(x_{ij})-y_{ij}\)^2,
\end{equation}
which is simply the average of the mean-squared error of each 
$\comp{g_i}{f}$ on $\zv_i$. 

An example of a neural network of the form $\comp{\gv}{\fbar}$ is illustrated 
in figure \ref{gnnet}. With reference to the figure, consider 
the problem of 
computing the derivative of $E(\comp{\gv}{\fbar},\z)$
with respect to a weight in the $i^{th}$ output network $g_i$. Denoting the 
weight by $w_i$ and recalling equation \eqref{gnferr}, we have
\begin{equation}
\label{ggrad}
\frac{\partial}{\partial w_i} E(\comp{\gv}{\fbar},\z) = 
\frac1n\frac{\partial}{\partial w_i}\frac1m\sum_{j=1}^m 
\(\comp{g_i}{f}(x_{ij}) - y_{ij}\)^2
\end{equation}
which is just $1/n$ times 
the derivative of the ordinary learning error \eqref{orderr}
of $\comp{g_i}{f}$
on sample $\zv_i=\((x_{i1},y_{i1}),\dots,(x_{im},y_{im})\)$ with respect to
the weight $w_i$. This can be computed using the standard backpropagation 
formula for the derivative \cite{Retal}.
Alternatively, if $w$ is a weight in the representation network $f$ then
\begin{equation}
\label{fgrad}
\frac{\partial}{\partial w} E(\comp{\gv}{\fbar},\z) =
\frac1n\sum_{i=1}^n \frac{\partial}{\partial w}\frac1m\sum_{j=1}^m
\(\comp{g_i}{f}(x_{ij})-y_{ij}\)^2
\end{equation}
which is simply the {\em average} of the derivatives of the ordinary 
learning errors over all the samples $(\zv_1,\dots,\zv_n)$ 
and hence can also be computed using the backpropagation formula.
\begin{figure}
\begin{center}
\leavevmode
\epsfxsize=8cm \epsfbox{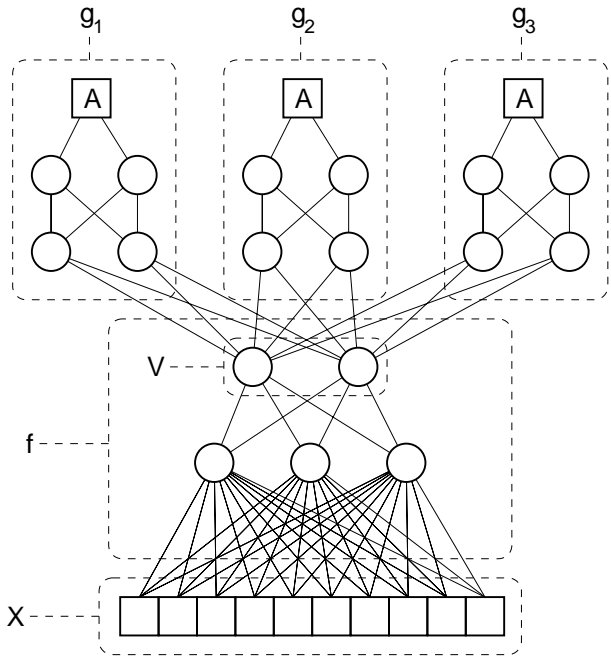}
\caption{A neural network for representation learning.\label{gnnet}}
\end{center}
\end{figure}

\section{Experiment: Learning translation invariance}
\label{expsec}
In this section the results of an 
experiment are reported in which a neural network like the one in figure \ref{gnnet}
was trained to perform a
very simple ``machine vision'' task where it had to learn certain 
{\em translationally invariant} Boolean functions.
All simulations reported here were performed on the 32 node CM5 at The South
Australian centre for Parallel Supercomputing.

The input space $X$ was viewed as a one-dimensional ``retina''
in which all the pixels could be either on (1) or off (0) (so in fact
$X=\{0,1\}^{10}$). However the network did not see all possible input vectors
during the course of its training, the only vectors with a non-zero probability
of appearing in the training set 
were those consisting of from one to four active adjacent pixels placed
somewhere in the input (wrapping at the edge was allowed).

The functions in the environment of the network consisted of
all possible {\em translationally invariant} Boolean functions over the 
input space (except the trivial ``constant 0'' and ``constant 1''
functions). The requirement of translation invariance means that the
environment consisted of just 14 different functions---all the Boolean
functions on four objects (of which there are $2^4=16$) less the two 
trivial ones. Thus the
environment was highly restricted, both in the 
number of different input vectors seen---40 out of a possible 1024---and
in the number of different functions to be learnt---14 out of a possible 
$2^{1024}$. $(n,m)$ samples were 
generated from this environment by firstly choosing $n$ functions 
(with replacement) uniformly from
the fourteen possible, and then choosing $m$ input vectors (with replacement 
again), for each function, uniformly from the 40 possible input vectors.

The architecture of the network was similar to the one shown in figure 
\ref{gnnet}, the only difference being that the output networks $g\in\G$ for
this experiment had only one hidden layer, not two. The network in 
figure \ref{gnnet} is for learning $(3,m)$ samples (it has $3$ output networks), in
general for learning $(n,m)$ samples the network will have $n$ output
networks.

The network was trained on $(n,m)$ samples with $n$ ranging from $1$ to
$21$ in steps of four and $m$ ranging from $1$ to $151$ in steps of
$10$. Conjugate-gradient descent was used with
exact line search with the gradients for each weight computed 
using the backpropagation algorithm according to the formulae \eqref{fgrad} and 
\eqref{ggrad}. Further details of the experimental procedure may be found in
\cite{thesis}.

Once the network had sucessfully learnt the $(n,m)$ sample its generalization
ability was tested on all $n$ functions in the training set.
In this case the generalisation
error (i.e {\em true} error---$E(\A(\z),\Pv)$) could be computed exactly by calculating the
network's output (for all $n$ functions) for each of the $40$ input vectors,
and comparing the result with the desired output.

In an ordinary
learning situtation the generalisation error of a 
network would be plotted as a function of $m$, the number of examples 
in the training set, resulting in a {\em learning curve}.
For representation learning there are two parameters $m$ and $n$ so the
learning curve 
becomes a {\em learning surface}.
Plots of the learning surface are shown in figure \ref{tplot}
for three independent simulations. 
All three cases support the theoretical 
result that the number of examples $m$ required for good generalisation
decreases with increasing $n$ ({\em cf} theorem \ref{thm1}). 

\begin{figure}
\begin{center}
\leavevmode
\epsfxsize=3.5in \epsfbox{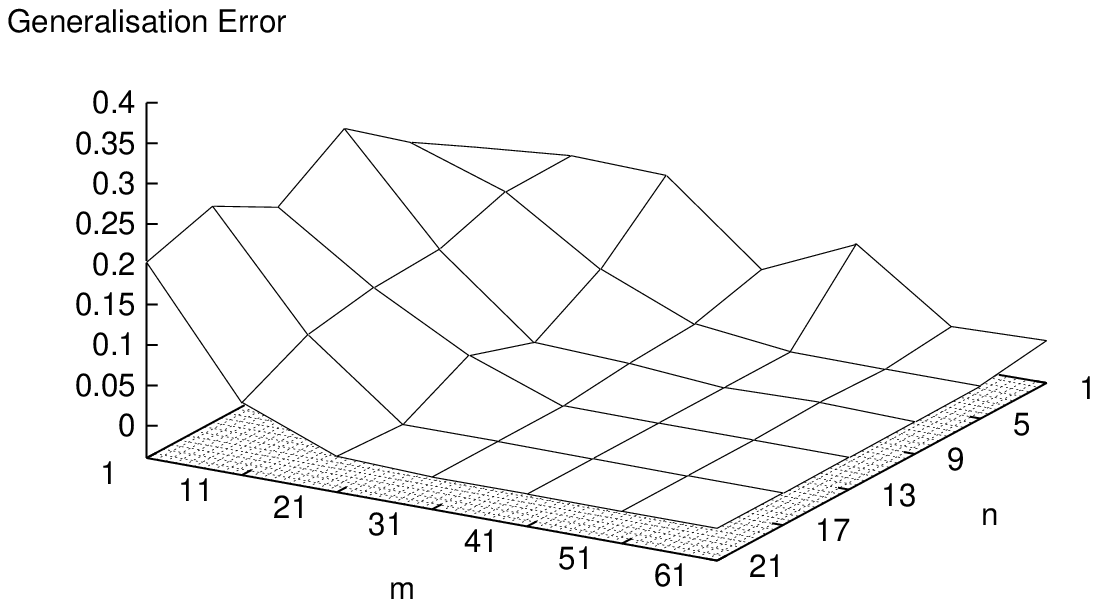}
\epsfxsize=3.5in \epsfbox{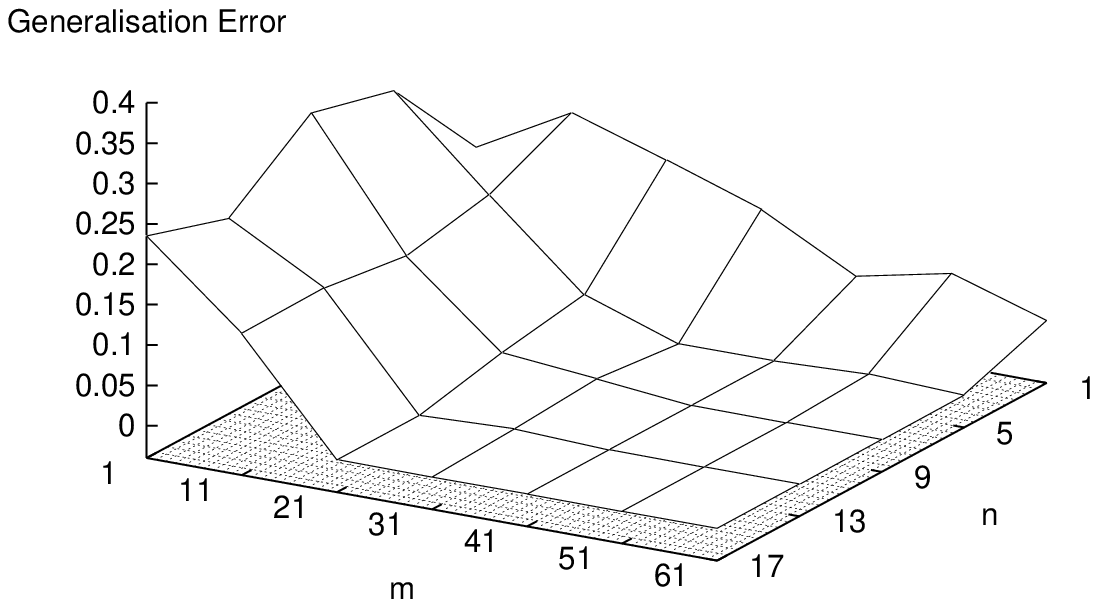}
\epsfxsize=3.5in \epsfbox{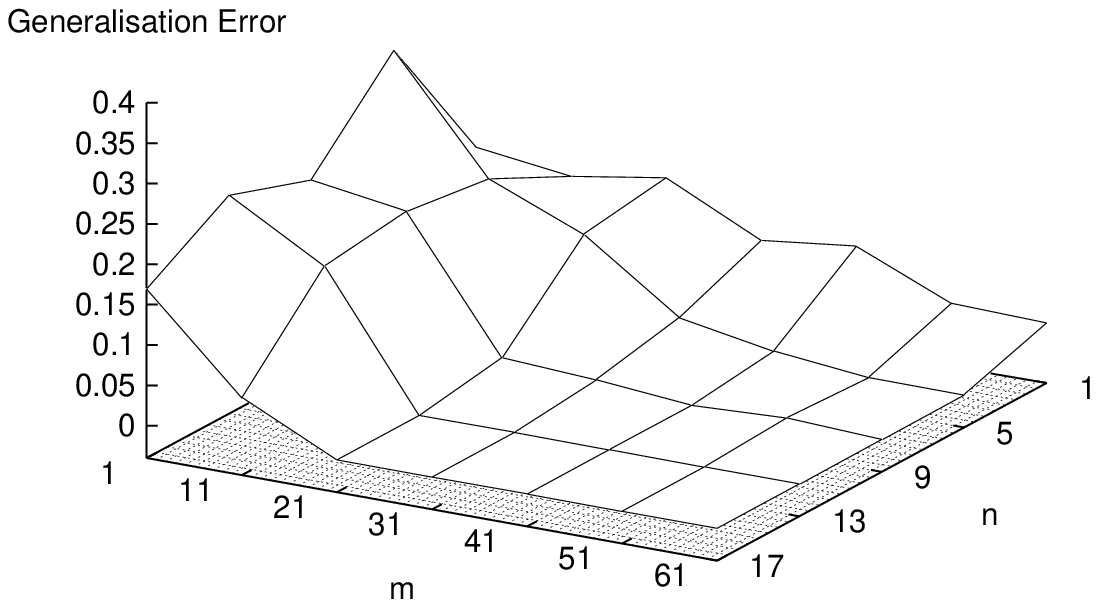}
\caption{Learning {\em surface} for three independent simulations.\label{tplot}}
\end{center}
\end{figure}

For $(n,m)$ samples that led to a generalisation error of less
than  $0.01$, the representation network $f$ was extracted and tested for 
its {\em true error}, where this is defined as in
equation \eqref{exploss} and in the current framework translates to
$$
E^*_\G(f,Q) = \frac1{560}\sum_{i=1}^{14}\inf_{g\in\G} \sum_{j=1}^{40}
\(\comp{g}{f}(x_j) - f_i(x_j)\)^2 
$$
where $x_1,\dots,x_{40}$ are the $40$ input vectors seen by the learner
and $f_1,\dots,f_{14}$
are all the functions in the environment. 
$E^*_\G(f,Q)$ measures how useful the representation $f$ is for learning all
functions in the environment, not just the ones used in generating the
$(n,m)$ sample $f$ was trained on. To measure $E^*_\G(f,Q)$, entire training
sets consisting of 40 input-output pairs were generated for each of the 14
functions in the environment and the training sets were learnt individually
by fixing $f$ and performing conjugate gradient descent on the
weights of $g$.
To be (nearly) certain
that a minimal solution had been found for each of the functions, learning
was started from $32$ different random weight initialisations for 
$g$ (this number was chosen so that the CM5 could
perform all the restarts in parallel) and the best result from all 32
recorded. For each
$(n,m)$ sample giving perfect generalisation, $E^*_\G(f,Q)$ was calculated and then
averaged over all $(n,m)$ samples with the same value of $n$, 
and finally averaged
over all three simulations, to give an indication
of the behaviour of $E^*_\G(f,Q)$ as a function of $n$. This is
plotted in figure \ref{E*}, along with the $L^\infty$ representation error
for the three simulations (i.e, the maximum error over all 14 functions and
all 40 examples and over all three simulations). Qualitatively the curves
support the theoretical conclusion that the representation error should
decrease with an increasing number of tasks $n$ in the $(n,m)$ sample.
However,  
note that the representation error is very small, even
when the representation is derived from learning only one function from the
environment. This can be explained as follows. 
For a representation to be a good one for learning in this
environment it must be translationally invariant and distinguish all
the four objects it sees (i.e.\ have different values on all four objects). 
For small values of $n$, to achieve perfect
generalisation the representation is forced to be at least approximately 
translationally invariant
and so half of the problem is already solved. However depending upon the
particular functions in the $(n,m)$ sample the representation may not have
to distinguish all four objects, for example it may map two objects
to the same element of $V$ if none of the functions
in the sample distinguish those objects (a function distinguishes two
objects if it has a different value on those two objects). However, because the
representation network is continuous it is very unlikely that it will
map different objects to
{\em exactly} the same element of $V$---there will always be slight
differences. When the representation is used to learn a function that does
distinguish the objects mapped to nearly the same element of $V$,
often an output network $g$ with sufficiently large weights can be found to
amplify this difference and produce a function with small error. This is why
a representation that is simply translationally invariant does quite well in
general. This argument is supported by a plot in figure \ref{outplot}
of the representation's behaviour for minimum-sized samples leading to
a generalisation error of less than $0.01$ for $n=1,5,13,17$.
The four different symbols marking the points in the
plots correspond to the four different input objects. 
For the $(1,111)$ plot the three and four pixel objects are well separated by the
representation while the one and two pixel objects are not, except that a 
closer look reveals that
there is a slight separation between the representation's output for the one
and two pixel objects. This separation can be exploited to learn a function
that at least partially 
distinguishes the two objects. Note that the representation's behaviour
improves dramatically with increasing $n$, in that all four
objects become well separated and the variation in the
representation's output for individual objects decreases. 
This improvement manifests itself in superior {\em learning curves} for
learning using a representation from a high $n$ $(n,m)$-sample, although it
is not necessarily reflected 
in the representation's error because
of the use of the infimum over all $g\in\G$ in the definition of that error.

\begin{figure}
\begin{center}
\leavevmode
\epsfxsize=1.6in \epsfbox{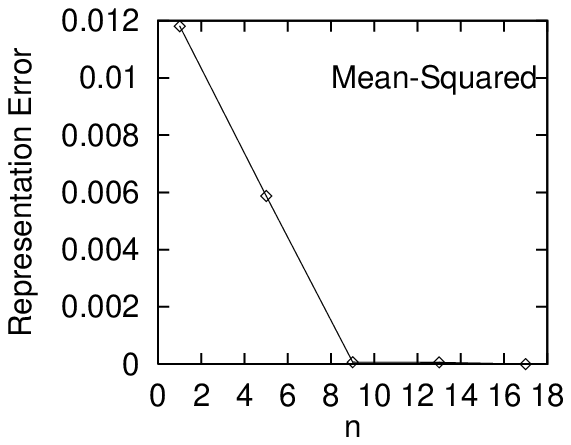}
\epsfxsize=1.6in \epsfbox{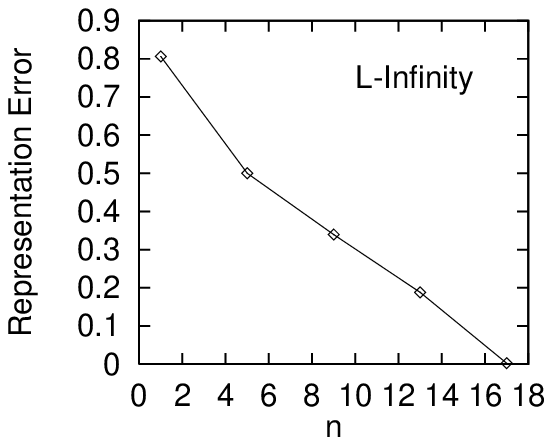}
\caption{Mean-Squared and $L^\infty$ representation error curves.\label{E*}}
\end{center}
\end{figure}
\begin{figure}
\begin{center}
\leavevmode
\epsfxsize=1.6in\epsfbox{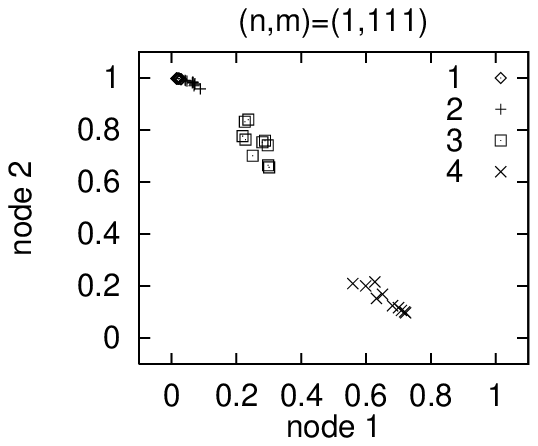}
\epsfxsize=1.6in\epsfbox{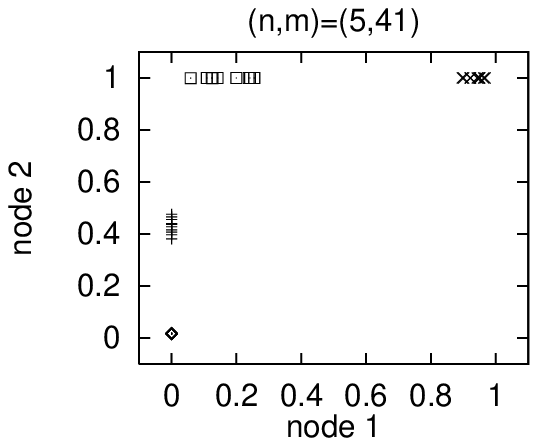}
\epsfxsize=1.6in\epsfbox{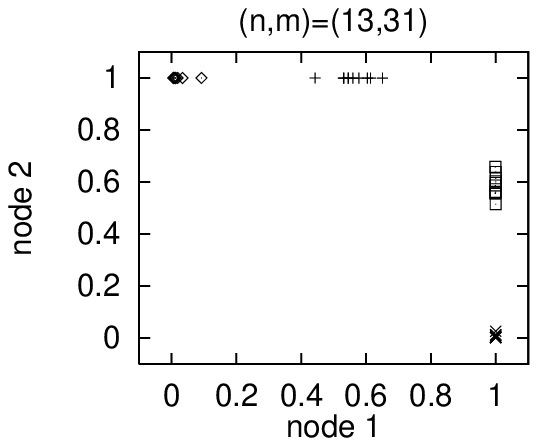}
\epsfxsize=1.6in\epsfbox{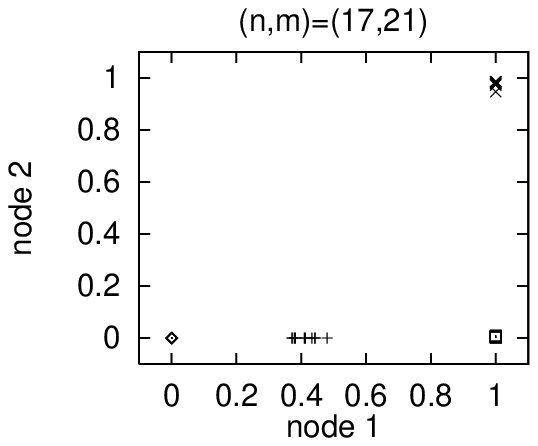}
\caption{Plots of the output of a representation generated from
the indicated $(n,m)$ sample. \label{outplot}}
\end{center}
\end{figure}

\subsection{Representation vs. no representation.}

As well as reducing the sampling burden for the $n$ tasks in the training
set, a representation learnt on sufficiently many tasks should be good for
learning novel tasks and should greatly reduce the number of examples
required of new tasks.
This was experimentally verified by taking a
representation $f$ known to be perfect for the environment above
and using it to
learn all the functions in the environment in turn. Hence the hypothesis
space of the learner was 
$\comp{\G}{f}$, rather than the full space
$\comp{\G}{\F}$. All the functions in the environment
were also learnt using the full space. The learning curves (i.e. the
generalisation error as a function of the number of examples in the training
set) were calculated for all 14 functions in each case. The learning
curves for all the functions were very similar and two are
plotted in figure 
\ref{repcomp}, for learning with a representation (Gof in the graphs) 
and without (GoF). 
These curves are the average of 32
different simulations obtained by using different random starting points 
for the weights in $g$ (and $f$ when using the full space to learn). 
Learning with a
good representation is clearly far superior to learning without.

\begin{figure}
\begin{center}
\leavevmode
\epsfxsize=1.6in\epsfbox{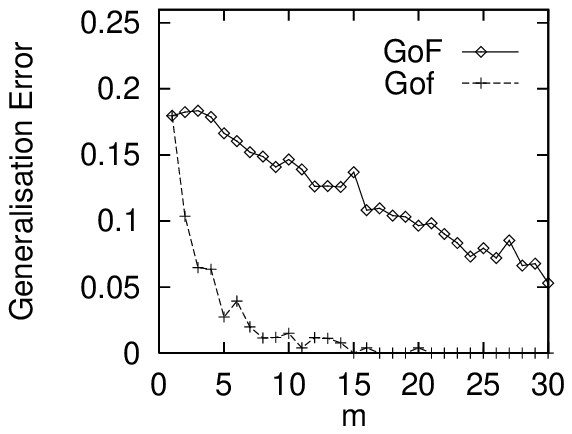}
\epsfxsize=1.6in\epsfbox{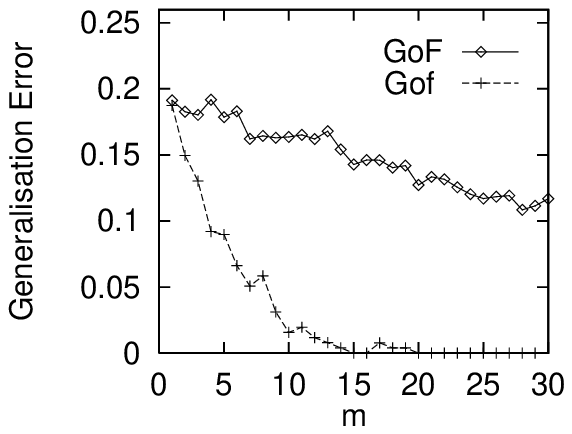}
\caption{Two typical learning curves for learning with a representation (Gof) 
vs. learning without (GoF).\label{repcomp}}
\end{center}
\end{figure}

\appendix
\section{Sketch proofs of theorems 1 and 2}
\label{proofapp}
\begin{defn}
\label{yuck}
{\rm
Let $\H_1,\dots,\H_n$ be $n$ sets of functions mapping $Z$ into $[0,M]$.
For any $h_1\in\H_1,\dots,h_n\in\H_n$, let 
$h_1\oplus\dots\oplus h_n$ or simply $\hv$ denote the map $\zv \mapsto
1/n\sum_{i=1}^n h_i(z_i)$ for all $\zv=(z_1,\dots,z_n)\in Z^n$. Let
$\H_1\oplus\dots\oplus\H_n$ denote the set of all such functions.
Given $m$ elements of $Z^n$, or equivalently an element of $\Zmn$,
$\z=(\zv_1,\dots,\zv_m)$, let $\E{\hv}{\z} =
1/m\sum_{i=1}^m \hv(\zv_i)$. For any product probebility measure
$\Pv=P_1\times\dots\times P_n$ on $\Zn$ let $\E{\hv}{\Pv}= \int_\Zn
\hv(\zv)\,d\Pv(\zv)$. For any set $\H\subseteq\H_1\oplus\dots\oplus\H_n$
define the pseudo-metric $d_\Pv$ and
$\ep$-capacity $\C(\ep,\H)$ as in \eqref{pmetric} and \eqref{capacity}.}
\end{defn}

The following lemma is a generalisation of a similar result (theorem 3) for
ordinary learning in
\cite{Haussler}, which is in turn derived from results in \cite{Pollard}. It
is proved in \cite{thesis} where it is called {\em the fundamental theorem}.
The definition of permissibility is given in appendix \ref{permapp}.
\begin{lem}
\label{fundthm}
Let $\H\subseteq\H_1\oplus\dots\oplus\H_n$ be a permissible set of functions
mapping $\Zn$ into $[0,M]$. Let $\z\in\Zmn$ be generated by
$m\geq\frac{2M}{\alpha^2\nu}$ independent trials
from $\Zn$ according to some product probability measure
$\Pv=P_1\times\dots\times P_n$. For all $\nu>0$,
$0<\alpha<1$, 
\begin{multline*}
\Pr\left\{\z\in\Zmn\colon \exists\hv\in\H\colon
\d{\E{\hv}{\z}}{\E{\hv}{\Pv}} > \alpha \right\} \\
\leq 4 \C\(\alpha\nu/8,\H\) \exp\left(-\alpha^2\nu nm/8M\right).
\end{multline*}
\end{lem}

\subsection{Proof sketch of theorem \ref{thm1}}
Let $l_\comp{\G^n}{\Fbar}$ denote the set of all functions
$\zv\mapsto1/n\sum_{i=1}^n l_\comp{g_i}{f}(z_i)$ where $g_1,\dots,g_n\in\G,
f\in\F$ and $\zv\in \Zn$, and let $l_\comp{\gv}{\fbar}$ denote an individual
such map. Recall that in theorem \ref{thm1} the learner $\A$ maps
$(n,m)$-samples into $\comp{\G^n}{\Fbar}$ and note from definition
\ref{yuck} and equation \eqref{egnferr} that
$E(\comp{\gv}{\fbar},\z) = \E{l_\comp{\gv}{\fbar}}{\z^T}$ where $T$ denotes
matrix transposition. This gives, 

\begin{align*}
  \Pr&\left\{\z\in\Znm \colon \d{E(\A(\z),\z)}{E(\A(\z),\Pv)}>\alpha\right\}  \\
  \begin{split}
    &\leq\Pr\left\{\z\in\Znm \colon\exists\comp{\gv}{\fbar}\in\comp{\G^n}{\Fbar}\colon\right.\\
    &\qquad\qquad\qquad\left.\d{E(\comp{\gv}{\fbar},\z)}{E(\comp{\gv}{\fbar},\Pv)}> \alpha\right\}
  \end{split}   \\
  \begin{split}
    &=\Pr\left\{\z\in\Zmn \colon \exists l_\comp{\gv}{\fbar}\in l_\comp{\G^n}{\Fbar} \colon\right. \\
    &\qquad\qquad\qquad\left.\d{\E{l_\comp{\gv}{\fbar}}{\z}}{\E{l_\comp{\gv}{\fbar}}{\Pv}} >\alpha\right\}
  \end{split} \\
  &\leq 4 \C\(\alpha\nu/8,l_\comp{\G^n}{\Fbar}\) \exp\left(-\alpha^2\nu nm/8M\right)
\end{align*}
The probability measure on $\Znm$ is
$P_1^m\times\dots\times P_n^m$ while on $\Zmn$ it is
$(P_1\times\dots\times P_n)^m$. With these measures the map $\z\mapsto\z^T$
is measure preserving, hence the
equality above.
The last inequality follows from lemma \ref{fundthm} noting that
$l_\comp{\G^n}{\Fbar}\subseteq l_\comp{\G}{\F}\oplus\dots\oplus
l_\comp{\G}{\F}$ and that $l_\comp{\G^n}{\Fbar}$ is permissible by the assumption
of f-permissibility of $\{l_\comp{\G}{f}\colon f\in\F\}$ and lemma
\ref{permlem} (permissibility of $[H^n]_\sigma$ in that lemma). 

Recalling definition \ref{metricdef}, it can be shown (see \cite{thesis},
appendix C) that
$$
\C\(\alpha\nu/8,l_\comp{\G^n}{\Fbar}\)\leq \C(\ep_1,l_\G)^n \C^*_{l_\G}(\ep_2,\F)
$$
where $\ep_1+\ep_2=\alpha\nu/8$.
Substituting this into the expression above and setting the result less than
$\delta$ gives theorem \ref{thm1} $\Box$

\subsection{Proof sketch of theorem \ref{thm2}}

To prove theorem \ref{thm2} note that in the $(n,m)$-sampling process a list
of probability measures $\Pv=(P_1,\dots,P_n)$ is implicitly 
generated in addition to the $(n,m)$-sample $\z$. Defining
$E^*_\G(f,\Pv)=1/n\sum_{i=1}^n\inf_{g\in\G}\E{l_\comp{g}{f}}{P_i}$
and using the triangle inequality on $d_\nu$, if
\begin{equation}
\label{ineq1}
\Pr\left\{(\z,\Pv)\colon
\d{E^*_\G(\A(\z),\z)}{E^*_\G(\A(\z),\Pv)} 
> \frac\alpha2 \right\} \leq \frac\delta2,
\end{equation}
and
\begin{equation}
\label{ineq2}
\Pr\left\{(\z,\Pv)\colon
\d{E^*_\G(\A(\z),\Pv)}{E^*_\G(\A(\z),Q)} 
> \frac\alpha2 \right\} \leq \frac\delta2,
\end{equation}
then
$$
\Pr\left\{\z\colon \d{E^*_\G(\A(\z),\z)}{E^*_\G(\A(\z),Q)} 
> \alpha \right\} \leq \delta.
$$
Inequality \eqref{ineq1} can be bounded using essentially 
the same techniques as theorem
\ref{thm1}, giving 
$$
m \geq \frac{32M}{\alpha^2\nu}\left[\ln \C\(\ep_1,l_\G\) + \frac1n
\ln\frac{8\C^*_{l_\G}\(\ep_2,\F\)}{\delta}\right],
$$
where $\ep_1+\ep_2 = \frac{\alpha\nu}{16}$.
Note that the probability in inequality \eqref{ineq2} is less than or equal to
\begin{equation}
\label{temp2}
\Pr\left\{\Pv\colon  \exists f\in\F\colon 
\d{E^*_\G(f,\Pv)}{E^*_\G(f,Q)} >\frac\alpha2 \right\}
\end{equation}
Now, for each $f\in\F$ define $l^*_f\colon\P\to [0,M]$ by
$l^*_f(P)=\inf_{g\in\G} \E{l_\comp{g}{f}}{P}$ and let $l^*_\F$ denote the
set of all such functions. Note that the expectation of $l^*_f$ with
respect to $\Pv=(P_1,\dots,P_n)$ satisfies $\E{l^*_f}{\Pv}=E^*_\G(f,\Pv)$ and
similarly $\E{l^*_f}{Q}= E^*_\G(f,Q)$. Hence \eqref{temp2} is equal to
\begin{equation}
\label{temp3}
\Pr\left\{\Pv\in \P^n\colon  \exists l^*_f\in l^*_\F\colon 
\d{\E{l^*_f}{\Pv}}{\E{l^*_f}{Q}} >\frac\alpha2 \right\}.
\end{equation}
For any probability measure $Q$ on $\P$ define the pseudo-metric 
$d_Q$ on $l^*_\F$ by
$$
d_Q(l^*_f, l^*_{f'}) = \int_\P |l^*_f(P) - l^*_{f'}(P)|\, dQ(P)
$$
and let $C(\ep,l^*_\F)$ be the corresponding $\ep$-capacity. Lemma
\ref{fundthm} with $n=1$ 
can now be used to show that \eqref{temp3} is less than or equal to
$4 C(\alpha\nu/16,l^*_\F)\exp(-\alpha^2\nu n/32 M)$ (permissibility of
$l^*_\F$ is guaranteed by f-permissibility of $\{l_\comp{\G}{f}\colon
f\in\F\}$---see lemma \ref{permlem} (permissibility of $H^*$ in that lemma)).
For any probability measure $Q$ on $\P$, let $Q_Z$ be the measure 
on $Z$ defined by $Q_Z(S) = \int_\P P(S)\,dQ(P)$ for any $S$ in the
$\sigma$-algebra on $Z$. It is then possible to show that
$d_Q(l^*_f,l^*_{f'}) \leq d^*_{[Q_Z,l_\G]}(f,f')$ (recall definition
\ref{metricdef}) and so $\C(\ep,l^*_\F) \leq \C^*_{l_\G}(\ep,\F)$ which
gives the bound on $n$ in theorem \ref{thm2} $\Box$

\section{F-permissibility and measurability}
\label{permapp}
In this section all the measurability conditions required to ensure theorems
\ref{thm1} and \ref{thm2} carry through are given. They are presented
without proof---the interested reader is referred to \cite{thesis} for the
details.
\begin{defn}
\label{measdef}
{\rm
Given any set of functions $\H\colon Z\to [0,M]$, where $Z$ is any set, 
let $\sigma_\H$ be the
$\sigma$-algebra on $Z$ generated by all inverse images under functions in $\H$
of all open balls (under the usual topology on $\R$) in $[0,M]$. 
Let $\P_\H$ denote the set of all probability measures on $\sigma_\H$.}
\end{defn}
This definition ensures measurability of any function $h\in\H$ with respect
to any measure $P\in\P_\H$. Use this definition anywhere in the rest of the
paper where a set needs a $\sigma$-algebra or a probability measure

The following two definitions are taken (with minor modifications) from
\cite{Pollard}.
\begin{defn}
\label{index}
{\rm
$\H\colon Z\to [0,M]$
is {\em indexed} by the set $T$ if
there exists a function $f\colon Z\times T\to [0,M]$ such that 
$$
\H = \left\{f(\,\cdot\,,t)\colon t\in T\right\}.
$$ }
\end{defn}
\begin{defn}
\label{permissible}
{\rm
$\H$ is {\em permissible} if it can be indexed by a set $T$ 
such that
$T$ is an {\em analytic} subset of a Polish space $\Tbar$, and
the function $f\colon Z\times T\to [0,M]$ indexing $\H$ by $T$ is measurable
with respect to the product $\sigma$-algebra $\sigma_\H\otimes \sigma(T)$, 
where
$\sigma(T)$ is the Borel $\sigma$-algebra induced by the topology on $T$.}
\end{defn}

For representation learning the concept of
permissibility must be extended to cover {\em hypothesis space families}, that
is, sets $H=\{\H\}$ where each $\H\in H$ is itself a set of functions
mapping $Z$ into $[0,M]$. Let $H_\sigma = \{h\colon h\in\H\colon\H\in H\}$.

\begin{defn}
\label{fpermdef}
{\rm
$H$ is {\em f-permissible}
if there exist sets
$S$ and $T$ that are analytic subsets of Polish spaces $\Sbar$ and $\Tbar$
respectively, and 
a function $f\colon Z\times T\times S\to [0,M]$, measurable with respect to 
$\sigma_{H_\sigma}\otimes\sigma(T)\otimes\sigma(S)$, such that 
$$
H = \bigl\{\{f(\,\cdot\,,t,s)\colon  t\in T\}\colon  s\in S\bigr\}.
$$}
\end{defn}

\begin{defn}
\label{blah}
{\rm
For any hypothesis space family $H$,
define $H^n=\{\H\oplus\dots\oplus\H\colon  \H\in H\}$.
For all $h\in H_\sigma$ define $\hbar\colon \P_{H_\sigma}\to[0,M]$ by 
$\hbar(P) = \E{h}{P}$. Let $\overline{H}_\sigma = \{\hbar\colon h\in
H_\sigma\}$ and for all $\H\in H$ let $\Hbar = \{\hbar\colon h\in\H\}$.
For all $\H\in H$ define $\H^*\colon \P_{H_\sigma}\to [0,M]$ by 
$\H^*(P) = \inf_{h\in\H} \hbar(P)$. Let $H^* = \{\H^*\colon \H\in H\}$.
For any probability measure $Q\in \P_{\overline{H}_\sigma}$, define the
probability measure $Q_Z$ on $Z$ by $Q_Z(S) = \int_{\P_{H_\sigma}}
P(S)\,dQ(P)$, for all $S\in \sigma_{H_\sigma}$.}
\end{defn}
Note that if $H=\{l_\comp{\G}{f}\colon f\in\F\}$ then $H^*=l^*_\F$ and
$[H^n]_\sigma=l_\comp{\G^n}{\Fbar}$. The f-permissibility of a hypothesis
space family $H$ is important for the permissibility and
measurability conditions it implies on $H^*,[H^n]_\sigma$, etc, as given in
the following lemma. Nearly all the results in this lemma are needed to
ensure theorems \ref{thm1} and \ref{thm2} hold.
\begin{lem}
\label{permlem}
Let $H$ be a family of hypothesis spaces mapping $Z$ into
$[0,M]$. Take the $\sigma$-algebra on $Z$ to be $\sigma_{H_\sigma}$, the set
of probability measures on $Z$ to be $\P_{H_\sigma}$ and the $\sigma$-algebra
on $\P_{H_\sigma}$ to be $\sigma_{\overline{H}_\sigma}$.
With these choices, if $H$ is f-permissible then
\begin{enumerate}
\item\label{1} $H_\sigma$, $[H^n]_\sigma$, $\overline{H}_\sigma$ and $H^*$
               are all permissible. 
\item\label{2} $\Hbar$ and $\H$ are permissible for all $\H\in H$.
\item\label{3} $H^n$ is f-permissible.
\item\label{4} $\H^*$ is measurable for all $\H\in H$.
\item\label{5} For all $Q\in \P_{\overline{H}_\sigma}$, $Q_Z\in \P_{H_\sigma}$.
\end{enumerate}
\end{lem}

{\large\bf Acknowledgements}

I would like to thank several anonymous referees for their helpful comments
on the original version of this paper.

\bibliographystyle{abbrv}
\bibliography{/home/staff/jon/bib/bib}
}
\end{document}

%% file: mathdefs.tex
\renewcommand{\Box}{\qedhere}
\newcommand{\mathbold}[1]{\mbox{\boldmath $\bf#1$}}

\renewcommand{\glossary}[2]{}
\def\addcontentsline#1#2#3{} 

\newtheorem{thm}{Theorem}

\newtheorem{lem}[thm]{Lemma}

\newtheorem{defn}{Definition}

\newcommand{\comp}[2]{{#1\circ#2}}

\renewcommand{\d}[2]{d_\nu\left(#1,#2\right)}

\newcommand{\rangel}[1]{\rangle_{#1}}
\newcommand{\E}[2]{\langle #1\rangel{#2}}

\renewcommand{\H}{{\cal H}}
\newcommand{\XY}{{X\times Y}}

\newcommand{\de}{=}
\renewcommand{\P}{{\cal P}}
\newcommand{\Zm}{{Z^m}}
\newcommand{\Zn}{{Z^n}}
\newcommand{\Zmn}{{Z^{(m,n)}}}
\newcommand{\Znm}{{Z^{(n,m)}}}

\newcommand{\z}{{\mathbold{z}}}

\newcommand{\w}{{\bf w}}

\newcommand{\be}{\begin{equation*}}

\newcommand{\ep}{{\varepsilon}}

\newcommand{\A}{{\cal A}}
\newcommand{\C}{{\cal C}}
\newcommand{\N}{{\cal N}}

\newcommand{\F}{{\cal F}}
\newcommand{\G}{{\cal G}}

\renewcommand{\(}{\left(}
\renewcommand{\)}{\right)}

\newcommand{\Pv}{{\vec{P}}}

\newcommand{\Fbar}{{\overline\F}}

\renewcommand{\hbar}{{\overline h}}

\newcommand{\fbar}{{\bar f}}

\newcommand{\zv}{{\vec{z}}}
\newcommand{\gv}{{\vec{g}}}

\newcommand{\hv}{{\vec h}}

\newcommand{\pr}[3]{{{#1}_1#2\dots#2{#1}_{#3}}}
\newcommand{\iton}{{1\leq i\leq n}}

\newcommand{\R}{{\mathbb R}}

\newcommand{\Tbar}{{\overline T}}
\newcommand{\Sbar}{{\overline S}}

\newcommand{\Hbar}{{\overline \H}}

\newcommand{\XVA}{X\stackrel{\F}{\rightarrow}V\stackrel{\G}{\rightarrow}A}

%% file: repcolt.bbl
\begin{thebibliography}{1}

\bibitem{thesis}
J.~Baxter.
\newblock {\em Learning Internal Representations}.
\newblock PhD thesis, Department of Mathematics and Statistics, The Flinders
  University of South Australia, 1995.
\newblock Draft copy in Neuroprose Archive under
  ``/pub/neuroprose/Thesis/baxter.thesis.ps.Z''.

\bibitem{Getal}
S.~Geman, E.~Bienenstock, and R.~Doursat.
\newblock Neural networks and the bias/variance dilemma.
\newblock {\em Neural Comput.}, 4:1--58, 1992.

\bibitem{Haussler}
D.~Haussler.
\newblock Decision theoretic generalizations of the pac model for neural net
  and other learning applications.
\newblock {\em Inform. Comput.}, 100:78--150, 1992.

\bibitem{Pollard}
D.~Pollard.
\newblock {\em Convergence of Stochastic Processes}.
\newblock Springer-Verlag, New York, 1984.

\bibitem{Retal}
D.~Rumelhart, G.~Hinton, and R.~Williams.
\newblock Learning representations by back-propagating errors.
\newblock {\em Nature}, 323:533--536, 1986.

\end{thebibliography}
